\pdfoutput=1
\documentclass[11pt]{article}

\usepackage{EACL2023}

\usepackage{times}
\usepackage{latexsym}
\usepackage{arydshln}
\usepackage[T1]{fontenc}
\usepackage[utf8]{inputenc}

\usepackage{microtype}

\usepackage{inconsolata}
\usepackage{microtype}

\usepackage{booktabs}
\usepackage{color}
\usepackage{amsmath}

\usepackage{graphicx}
\usepackage{multirow}

\usepackage{stfloats}

\usepackage{color}
\definecolor{shun}{rgb}{0.1,0.1,0.9}

\title{Metaphor Detection with Effective Context Denoising}

\author{Shun Wang\textsuperscript{1}, Yucheng Li\textsuperscript{2}, Chenghua Lin\textsuperscript{1}\thanks{~~Corresponding author}~, Loïc Barrault\textsuperscript{1,3}, Frank Guerin\textsuperscript{2} \\
\textsuperscript{1}~Department of Computer Science, University of Sheffield, UK \\
\textsuperscript{2}~Department of Computer Science, University of Surrey, UK \\
\textsuperscript{3}~Meta AI \\
\texttt{\{swang209, c.lin\}@sheffield.ac.uk}\\
\texttt{\{yucheng.li, f.guerin\}@surrey.ac.uk}\\
\texttt{loicbarrault@meta.com}
}
\begin{document}
\maketitle
\begin{abstract}
We propose a novel RoBERTa-based model, RoPPT, which introduces a target-oriented parse tree structure in metaphor detection. Compared to existing models, RoPPT focuses on semantically relevant information and achieves the state-of-the-art on several main metaphor datasets. We also compare our approach against several popular denoising and pruning methods, demonstrating the effectiveness of our approach in context denoising. Our code and dataset can be found at \url{https://github.com/MajiBear000/RoPPT}.
\end{abstract}

\section{Introduction} \label{sec:introduction}

Metaphor is a pervasive linguistic device, which attracts attention from both the fields of psycholinguistics and computational linguistics due to the key role it plays in the cognitive and communicative functions of language \citep{wilks1978-spv2,lakoff1980metaphors, lakoff1993contemporary}. Linguistically, metaphor is defined as a figurative expression that uses one or several words to represent another concept given the context, rather than taking the literal meaning of the expression \citep{fass1991met}. 
For instance, in the sentence ``\textit{This project is such a \underline{headache}!}'', the contextual meaning of \textit{headache} is ``a thing or person that causes worry or trouble'', different from its literal meaning, ``a continuous pain in the head''.\textcolor{blue}{\footnote{\url{https://www.oxfordlearnersdictionaries.com}}}
%\textcolor{red}{[CL: where is the source of the meaning from? We may need to mention the source.]}

%For instance, in the sentence ``A \textit{human \underline{drinks} water, but a car \underline{drinks} gasoline.}'', the two  ``drinks'' have different meanings. The first  \textit{drinks} is a normal expression which only has a literal meaning. \textcolor{orange}{However, the second  ``drinks'' is a metaphor which extends  the meaning to a different manner of consumption. Therefore, the meaning of “drinks” is context-specific, and the metaphor expression depends on context.}

Metaphor detection is challenging, as it requires understanding the nuanced relationships between abstract concepts embodied by the metaphoric expression and its surrounding context. Recent studies on this direction show its potential in benefiting a wide range of NLP applications, including sentiment analysis \cite{Li2023TheSO}, metaphor generation \cite{Li2022CMGenAN,Li2022NominalMG} and mental healthcare \cite{abd2017analysing,gutierrez-etal-2017-using-mental-health-care}.
When modelling relevant context for metaphor detection, various strategies have been proposed. These range from using highly restricted forms of linguistic context such as subject-verb and verb-direct object word pairs~\cite{gutierrez-etal-2016-literal}, to a wider context accounting for a fixed window surrounding the target word~\cite{do-dinh-gurevych-2016-token,mao2018word}, and modelling the full sentential context~\cite{gao2018neural,choi2021melbert}. While it has been argued that modelling a wider context is beneficial~\cite{cheng-etal-2021-guiding}, it has also been noted that a wider context is likely to introduce noise into the representations, and hence hinder model's performance in metaphor detection~\cite{le2020multi}.

Some recent efforts~\cite{le2020multi,song-etal-2021-verb} attempt to improve context modelling by explicitly leveraging the syntactic structure (e.g., dependency parse tree) of a sentence in order to capture important context words, where the parse trees are typically encoded with graph convolutional neural networks. MelBERT \cite{choi2021melbert} employs a simple chunking method which separates sub-sentences by commas. The sub-sentence that contains a target word is then marked with a special token type, signalling its contextual importance to the target. 
However, these strategies are either difficult to apply to batch optimisation due to their tree-dependent encoding process, or have limited effectiveness for context denosing. 
%\textcolor{blue}{
For instance, the simple chunking mechanism misses the syntactic structure, and thus can neither determine the degree of importance of context words, nor connect information across different subsentences.

In this paper, we propose a novel metaphor detection model RoPPT: RoBERTa with Pruning on target-oriented Parse Tree. 
RoPPT introduces a \textit{flat}, \textit{target-oriented} tree structure by reshaping and pruning the ordinary parse trees to extract semantically relevant neighbours of a target word. The resulting tree representation allows the model to focus on syntactically relevant information of a target word, and ignore irrelevant parts despite their position. It thus retains more relevant context for metaphor detection.

%We evaluate our model against the state-of-the-art models for metaphor detection. 
Extensive experiments conducted on three public benchmark datasets (i.e., VUA, MOH-X, TroFi) show that RoPPT can significantly improve metaphor detection on all datasets against several popular denoising and pruning methods.
%, compared to baseline models which do not apply the target word oriented pruning process.
Our model also yields better or comparable performance to the state-of-the-art models \cite{choi2021melbert,song-etal-2021-verb} in Micro F1 measure. 
To further validate our approach, we conducted an additional investigation to assess the effect of sentence length on the performance of our model. Experimental results demonstrate a positive correlation between the increase in the performance of RoPPT and the length of the input sentence. 
% To further validate the effectiveness of our approach, we also investigated  a contrast test based on sentence length 
% %which shows our RoPPT beats SOTA on long samples.
% which shows a positive correlation between performance boost (of RoPPT over SOTA) and the sentence length.

%compare our approach against several popular denoising and pruning methods.

In summary,  our paper makes three contributions: 
(1) we propose a flat, target word-oriented tree structure by reshaping and pruning the ordinary parse trees to retain the most relevant context for a target word; 
(2) we propose RoPPT, a RoBERTa-based model which can effectively encode the target-oriented parse tree for metaphor detection, achieving state-of-the-art results on three bench mark datasets;
(3) we compare and evaluate a range of context denoising methods for metaphor detection, demonstrating the effectiveness of our proposed tree structure in context denoising.

\begin{figure*}[ht]
    \centering
    \includegraphics[width=1.0\textwidth]{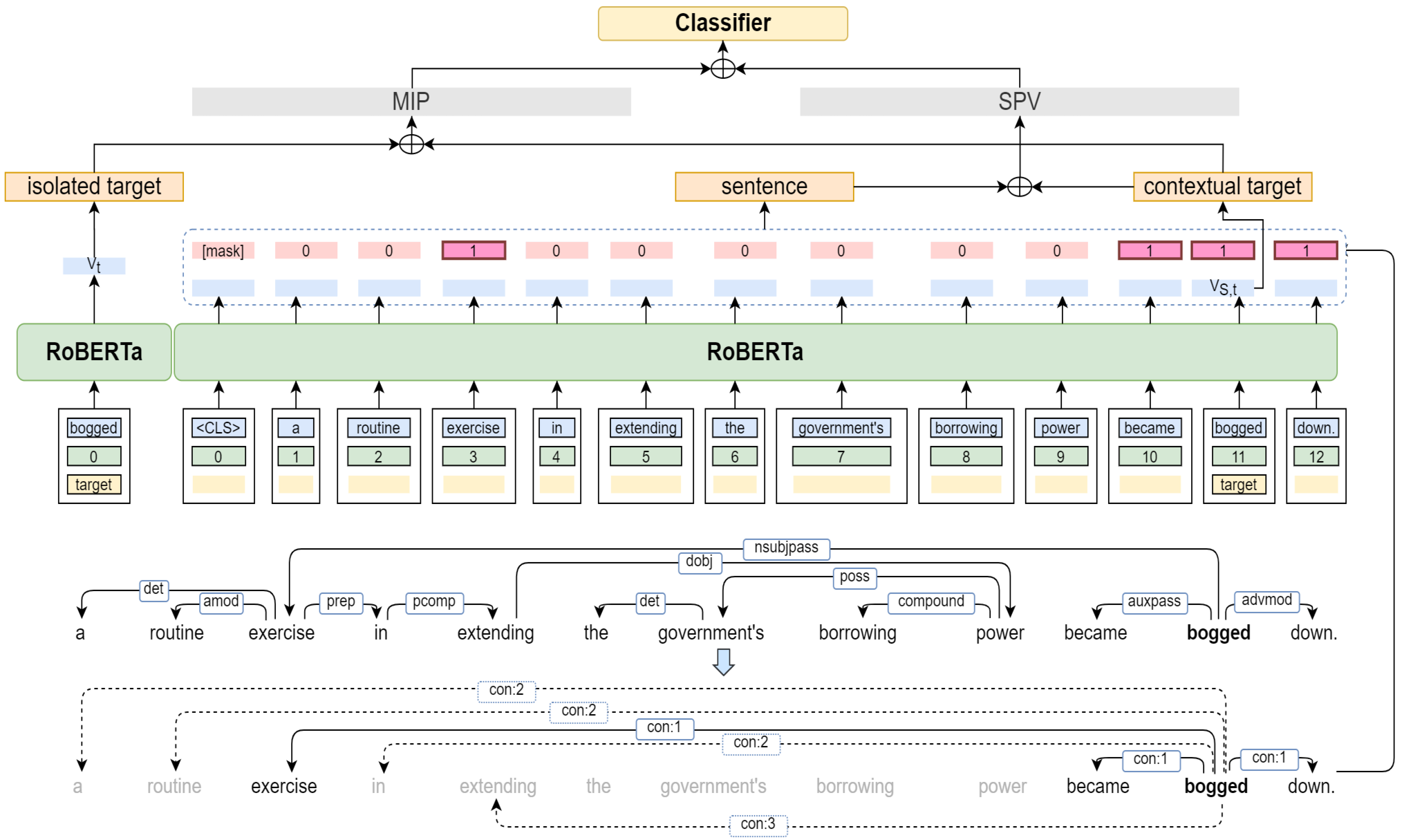}
    \caption{The overall framework of RoPPT. The parse tree of a sentence is reshaped to a target-oriented tree, and the context is pruned with a pre-set threshold. The sentence embedding is the average pooling result of hidden states for pruned context from RoBERTa. $\bigoplus$ denotes concatenation.}
    \label{fig:framework}
\end{figure*}

\section{Method}
\label{sec:methods}

The overall architecture of RoPPT is shown in Figure \ref{fig:framework}, which can be divided into two parts: a target-oriented parse tree pruning module and a RoBERTa \cite{liu2019roberta} contextual encoder. 
%We describe the  details of these two modules below. 
%In the pruning module, we obtain the target-oriented parse tree by rooting the target and removing irrelevant contexts. We then realize two linguistic theories and apply the pruned tree on the encoder to identify metaphors.

\subsection{Target-oriented Dependency Parse Tree}\label{parsetree}
Connecting target words with their most relevant context words 
is crucial for metaphor detection and comprehension.
%lies at the heart of the task.
While there have been attempts to employ dependency parse trees in graph convolutional neural networks to improve context modelling~\citep{wang2020relational}, it raises challenges of how to effectively encode and leverage such  syntactic structure information for transformer-based mask language models for metaphor detection.
%it is not uncommon that a sentence contains multiple metaphoric target words, which can easily lead to a target word being influenced by unrelated context words and relations
%As we analysed in \S\ref{sec:introduction}, it is sufficient for metaphor detection tasks to focus only on a small number of context words that are semantically close to the target word and involving irrelevant context might lead in noise. 
%Inspired by works of \citeauthor{wang2020relational},

We tackle this challenge by introducing a  target-oriented parse tree generated by three steps: \textbf{1}) reshape the original parse tree from existing parsers such as spaCy \citep{spacy2} and Biaffine \citep{dozat2016deep}; \textbf{2}) root the tree at the target word; \textbf{3}) prune the tree according to the distance 
%(set as hyperparameter \textit{neighbor range}) 
between leaves and root, coined as \textit{neighbor range}. The rationale behind is that the target word is the focus of the task rather than the original root. So the re-rooting allows us to focus on the connections between target words and their relevant context. The resulting flat, target-oriented tree structure also enables simple encoding process into the model.  
Figure \ref{fig:framework} shows an example of our reshaped tree, which retrains words with neighbor range $con=1$ to the root `bogged'. 

\subsection{RoBERTa-based Context Encoder}
\label{semantic encoder}
We employ two metaphor identification theories in our model, i.e., Metaphor Identification Procedure \cite[MIP]{steen2010method}  and Selectional Preference Violation \cite[SPV]{wilks1978-spv2}. In MIP, a metaphor is detected when there is a contrast between target word's contextual and literal meanings, whereas in SPV a metaphorical word is identified by the semantic difference from its surrounding words.
%in SPV it is the contrast  between the target word and its context. 
Therefore, we model three types of semantic representations for implementing MIP and SPV, 
i.e.,  the literal meaning and the contextual meaning of a target word, and the context meaning. 

Formally, given a sentence $S=(w_0, ..., w_n)$, we first employ the RoBERTa network to produce representations for each word.
\begin{equation}
    H=\mathrm{RoBERTa\_Enc}(\mathrm{emb}_{\text{cls}}, ..., \mathrm{emb}_{n})
\end{equation}
Here \textsc{cls} is a special token indicating the start of an input, $H=(h_{\text{cls}}, h_0, ..., h_n)$  the output hidden states, and $\mathrm{emb}_{i}$ the input embedding for word $w_i$. Specifically, $ \mathrm{emb}_{i} = \mathrm{emb}_w + \mathrm{emb}_{pos}$, 
% \begin{equation}
%     \mathrm{emb}_{i} = \mathrm{emb}_w + \mathrm{emb}_{pos} + \mathrm{emb}_{type} 
% \end{equation}
where $\mathrm{emb}_{w}$ is the word embedding, and $\mathrm{emb}_{pos}$ the position encoding.
%representing the position for each word, 
%and $\mathrm{emb}_{type}$ the type encoding for  distinguishing target words within the sentence. 

\noindent\textbf{Context denoising with the target-oriented parse tree.}~~When modelling sentence representation, existing works directly employed the \textsc{cls} embedding as a common practice~\citep{choi2021melbert, song2021verb}. In contrast, RoPPT employs the target-oriented parse tree to retain the most relevant context for a target word when computing the sentence embedding.  
%pruning irrelevant context and focus on closest neighbours of target word based on the parse tree obtained in \S~\ref{parsetree}
% obtain the sentence representation by focusing on target's syntactically related context on the target-oriented dependence tree 
%rather than directly employing the \textsc{cls} embedding as the sentence representation \citep{choi2021melbert, song2021verb}. 
Specifically, our sentence embedding is computed as follows.
\begin{equation}
\label{equa:sentence embedding}
    v_{S} = \frac{1}{n}\sum h_{i},\;  i \in \mathcal{C}_{n} % N_{\text{target}}
\end{equation}
Here $v_{S}$ is the sentence representation; $\mathcal{C}_{n}$ represents the $n$ neighbour words within the neighbour range of the target-oriented parse tree, and $h_i$ is the hidden state of $w_i$. 
In other words, we do average pooling on the most relevant context words as the sentence representation and ignore other words in the sentence.
%\footnote{We also experimented with another strategy which is to directly applies the dependency tree to filter the input sentence.  While fundamentally intuitive, we found that this approach  does not help improve the model performance empirically.}.
We also design an alternative strategy by directly masking the original input sentence to the encoder according to the pruned parse tree.   
%is another possible strategy to apply the pruned parse tree.
We denote this intuitive solution as RoPPT with Input Mask (\textbf{RoPPT\_IM}) and discuss the performance difference between these two variants in \S\ref{sec:results}.

Similar to \citet{choi2021melbert}, we use the hidden state of target word $w_t$ as the contextual target word embedding (i.e. $ v_{S,t} = h_{t}$), and the literal target word embedding $v_{t}$ is obtained by feeding the single target word $w_t$ to the RoBERTa network.
%in order to obtain its literal representation.   i.e. without contextual meaning.
\begin{equation}
\label{equa:literal target embedding}
    v_{t} = \mathrm{RoBERTa\_Enc}(\mathrm{emb}_t)
\end{equation}
%\textcolor{purple}{Then, we can represent SPV ($h_{\text{SPV}}$) by computing feature from the concatenation of $v_{S}$ and $v_{S,t}$, and MIP ($h_{\text{MIP}}$) by computing feature from the concatenation of $v_{S,t}$ and $v_{t}$.} 

We then model SPV ($h_{\text{SPV}}$) by concatenating the sentence embedding $v_{S}$  and contextual target embedding $v_{S,t}$, and MIP  ($h_{\text{MIP}}$) by concatenating the contextual and literal target embeddings $v_{t}$, followed by a MLP layer (i.e. $f_1{}(\cdot)$ and $f_2(\cdot)$).
%\textcolor{red}{The concatenation does not represent the contrasting relation per se, but provides all the needed information to the classification layer.}
% \begin{gather}
   % h_{\text{SPV}} = f_1([v_{S}, v_{S,t}]) ;\
   % h_{\text{MIP}} = f_2([v_{S,t}, v_{t}])
% \end{gather}
\begin{gather}
    h_{\text{SPV}} = f_1([v_{S}, v_{S,t}]) \\
    h_{\text{MIP}} = f_2([v_{S,t}, v_{t}])
\end{gather}

%The concatenation does not represent the contrasting relation per se, but provides all the needed information to the classification layer. 
%This approach has been proven effective by previous works using MIP and SPV \citep{mao2019end, choi2021melbert}. 

%\textcolor{shun}{where $f_1{}(\cdot)$ and $f_2(\cdot)$ denote a MLP layer to learn the contrast between two features.}
%, there are three semantic representations are obtained, where sentence embedding and contextual target embedding support SPV identification and contextual and literal target embedding support MIP identification. Specifically, we concatenate these three representation to get $h_{MIP}$ and $h_{SPV}$:

Finally, we combine two hidden vectors $h_{\text{MIP}}$ and $h_{\text{SPV}}$ to compute a prediction score $\hat{y}$, and use binary cross entropy loss to train the overall framework for metaphor prediction.
\begin{gather}
    \hat{y}=\sigma (W^{\top }[h_{\text{MIP}};h_{\text{SPV}}]+b) \\
    \mathcal{L}= -\sum_{i=1}^{N}[y_{i}\log\hat{y}_{i}+(1-y_{i})\log(1-\hat{y}_{i})]
\end{gather}

\section{Experimental Setup}
\label{experiment}
%\subsection{Dataset}
\noindent\textbf{Dataset.}~~We conduct experiments on four public benchmark datasets. 
\textbf{VUA-18} \citep{leong2018report} and \textbf{VUA-20} \citep{leong2020report} are the largest available datasets,
released in the metaphor detection shared tasks in 2018 and 2020.
VUA-20 extends VUA-18 with about 12K sentences for training set and 3.6K sentences for test and validation sets. 
%\textcolor{shun}{To facilitate comparison with baselines, the pre-processed version we used in this paper are released by \citeauthor{choi2021melbert} which is the SOTA on VUA.}
The \textbf{MOH-X} dataset is constructed by sampling  sentences from WordNet \cite{miller1998wordnet}. Only a single target verb in each sentence is annotated. The average sentence length is 8 tokens, the shortest of our three datasets. 
\textbf{TroFi} \cite{birke2006-trofi} consists of sentences from the 1987-89 Wall Street Journal Corpus 
%Release 1 
\cite{charniak2000-wsj8789}, with an average length of 28.3 tokens per sentence.

%\subsection{Baselines}
\noindent\textbf{Baselines.}~~\textbf{RoBERTa\_SEQ} \citep{leong2020report} is a fine-tuned RoBERTa sequence labeling model  for metaphor detection.
\textbf{MelBERT} \citep{choi2021melbert} realises MIP and SPV theories via a RoBERTa based model. 
\textbf{MrBERT} \citep{song2021verb} is the recent SOTA on verb metaphor detection based on BERT with  verb relations encoded. 
\\
\noindent\textbf{Hyperparameter.}~We set the hyperparameter neighbour range $con=4$  based on the validation set results. All the parser results are based on spaCy as it performs better than Biaffine empirically (see \S \ref{sec:results} for more discussion).
%Our experiment was run on TITAN RTX GPU, and it takes 50 minutes to run an epoch on VUA20 dataset.
%Runtime details are given in Table~\ref{tabel:experiment} in Appendix.

\begin{table}[]
\resizebox{\columnwidth}{!}{
    \centering \small
    \begin{tabular}{l|ccc|ccc}
    \toprule
           \multirow{2}{*}{Model}      & \multicolumn{3}{c|}{VUA18}  & \multicolumn{3}{c}{VUA20}    \\ \cline{2-7}
       & Prec & Rec  & F1            & Prec & Rec  & F1            \\ \hline
    RNN\_ELMo    & 71.6 & 73.6 & 72.6          & -    & -    & -             \\
    RoBERTa\_SEQ & 80.1 & 74.4 & 77.1          & 75.1 & 67.1 & 70.9          \\
    MrBERT       & 82.7 & 72.5 & 77.2          & -    & -    & -             \\
    MelBERT*     & 79.6 & 76.4 & 77.9          & 76.3 & 68.6 & 72.2          \\
    MelBERT      & 80.1 & 76.9 & 78.5          & 75.9 & 69.0 & 72.3         \\  \hline
    RoBERTa\_tree & 78.9        & 76.1        & 77.4                 & 74.8 & 68.6 & 71.6          \\
    RoChunk  & 76.6 & 80.0 & 78.2       & 73.9 & 70.0 & 71.9             \\
    RoWindow & 78.0 & 78.1 & 78.0       & 75.0 & 68.8 & 71.8          \\ \hdashline
    RoPPT\_IM    & 73.4 & 74.3 & 73.9          & 67.7 & 66.8 & 67.2          \\    
    RoPPT        & 80.0 & 78.2 & \textbf{79.1} & 75.9 & 70.0 & \textbf{72.8} \\ \bottomrule
    \end{tabular}
}
\caption{Performance comparison on \textbf{VUA} dataset (best is in \textbf{bold}). NB: * indicates the reproduced results of MelBERT using the original source code and setting of \cite{choi2021melbert}. RNN\_ELMo and MrBERT have no results on VUA20 in their original paper.
%, and $^\dag$ indicates the results which are reported by the prior publications
Popular denoising methods are also compared. \textbf{RoChunk} means chunk sentence by comma on RoBERTa input, \textbf{RoWindow} means denoising by a context window (size=4). \textbf{RoPPT\_IM} represent masking sentence before input to transformer encoder.}  
\label{tabel:VUA_result}
\end{table}

\begin{table}[t] \small
\resizebox{\columnwidth}{!}{
    \begin{tabular}{l|ccc|ccc} 
    \toprule
       \multirow{2}{*}{Models}     & \multicolumn{3}{c|}{TroFi}                    & \multicolumn{3}{c}{MOH-X} \\ \cline{2-7}
       & Prec          & Rec           & F1             & Prec  & Rec   & F1        \\ \hline
    RoBERTa\_SEQ  &53.6      & 70.1          & 60.7           & 80.6  & 77.7  & 78.7     \\
    DeepMet  & 53.7          & 72.9          & 61.7           & 79.9  & 76.5  & 77.9      \\
    MrBERT   & 53.8          & 75.0          & 62.7        & 75.9  & 84.1  & 79.8      \\
    MelBERT* & 53.1          & 73.2          & 61.6       & 78.0   & 79.5  & 78.8      \\
    MelBERT  & 53.4          & 74.1          & 62.0           & 79.3  & 79.7  & 79.2      \\ \hline
    RoBERTa\_tree  & 50.3           & 77.8                 & 61.1                  & 76.9         & 83.5         & 79.3        \\ \hdashline
    RoPPT    & 54.2 & 76.2 & \textbf{63.3}    & 77.0  & 83.5  & \textbf{80.1} \\ \bottomrule 
    \end{tabular}
    }
\caption{Performance comparison on TroFi and MOH-X datasets (NB: \textbf{bold} denotes the best result).}
\label{table:TroFI&MOH}
\end{table}

%\begin{table}[t] \small
%\resizebox{\columnwidth}{!}{
 %   \begin{tabular}{l|cc|c} 
  %  \toprule
   % Parsers  &  based-on    &    Precision  & F1 \\ \hline
    %spaCy   & RoBERTa  &   95    &   78.0 \\
    %Biaffine    & CNN     & 90    &   77.7 \\
    %\bottomrule 
    %\end{tabular}
    %}
%\caption{Impact of Parsers on the validation set.}
%\label{table:parsers}
%\end{table}

\section{Experimental Results}
\label{sec:results} 
\noindent\textbf{Overall results.}~~
Table~\ref{tabel:VUA_result} shows a comparison of the performance of our models against the baseline models on VUA18 and VUA20, respectively.  It is clear that our RoPPT outperforms all baselines on VUA18 and VUA20, including the state-of-the-art model MelBERT. A two-tailed $t$-test was conducted based on 10 paired results from RoPPT and the strongest baseline MelBERT* on both VUA-18 ($p=0.014$) and VUA-20 ($p=0.019$). 

We also compared our method against several common denoising strategies. The results show that our tree-based denoising method is more effective than other popular denoising approaches such as RoChunk and RoWindow, which are sequence-based methods. We also apply our target-oriented tree to RoBERTa\_SEQ, denoted as the RoBERTa\_tree model. The improvement of RoBERTa\_tree over RoBERTa\_SEQ on two VUA datasets (i.e. 0.3\% and 0.7\%) further demonstrates the utility of our tree-based denoising method. 
%effective not only on MelBERT but independent from the frameworks.   
%\textcolor{red}{[Important: (1) you need to discuss the effectiveness our parse tree based on the results reported in Table 1; and (2) compare with other denoising methods]}

%\noindent\textbf{Generalization.}~~
Following the setup of \citet{choi2021melbert}, we also conducted a zero-shot transfer learning experiment shown in Table~\ref{table:TroFI&MOH}. Specifically, our model is trained on the training set of VUA20 and directly tested on the entire Trofi and MOH-X datasets. This is intended to test the generalisation power of trained models. RoPPT shows the best performance on both datasets (significant test on RoPPT against MelBERT*: TroFi $p=0.0001$; MOH-X $p=0.021$;  we cannot compare with MrBERT as the code is unavailable).
It can be observed that our model gives a larger margin of improvement over the baselines on TroFi (i.e., 1.3\% gain over MelBERT and 0.6\%  over MrBERT) than MoH-X (i.e., 0.9\% gain over MelBRTT and 0.3\%  over MrBERT). 

\noindent\textbf{Model performance vs. Sentence length.}~~As the averaged sentences length of TroFi (28.3 tokens) is significantly longer than that of MoH-X (8 tokens), it is worth investigating whether our model gives more performance boost on data with longer context as it is likely to be noisier.
To verify this hypothesis, 
% we conduct a experiment below.
% \noindent\textbf{Model performance vs. Sentence length.}~~
we evaluated the performance boost of our RoPPT against the SOTA baseline MelBERT.  
Table~\ref{table:Length_Study} shows the results of VUA18 with the testset splitted into 3 different groups based on sentence length. The results demonstrate a clear positive correlation between performance boost and sentence length. %(similar patterns can also be observed in VUA20 but we cannot include the results due to space limit). 
%\textcolor{shun}{Considering TroFi proved to be more challenging as sentences can be as long as 118 tokens, the performance is improved by a relatively large margin. Due to the less complex examples of MOH-X, the performance of our method benefits less from the de-noising parse tree, which still exceeds MelBERT by F1 score of 0.9.}

\noindent\textbf{Impact of Parsers.}~~
We also investigated how the choice of parsers impacts the metaphor detection performance of our model. Specifically, we tested two parsers for constructing the target-oriented dependency parse trees, namely, the CNN-based parser Biaffine and the RoBERTa-based parser spaCy. When tested on the validation set, our model achieves 78.0\% with spaCy and 77.7\% with Biaffine in F1 for metaphor detection, respectively. This shows that the impact of the parse choice is relatively small for our model.

\noindent\textbf{Case Studies.}~~
RoPPT shows its strength in the following example with the target word far away from its subject, which is correctly labeled by RoPPT but incorrectly by baseline models. For the instance with metaphorical target word \textit{bogged}, "\textit{a routine exercise in extending the government's borrowing power to \$3.1 thousand billion became \underline{bogged} down.}", the target word \textit{bogged} is separated from its subject by a long phrase, which causes baselines (including MelBERT) to fail to detect the metaphor. Thanks to the parse tree, RoPPT links  \textit{exercise} directly to the target and produces the right label. 
%\textcolor{red}{[If no space, can you show the parse tree in Appendix please? This is also suggested by Loic.]}
% Our model can prune the part after '\textit{born}' well to achieve the purpose of denoising.  
%(2) RoPPT is able to solve false positive examples where a target verb's neighbours mislead baselines. % and its real subject is in the distance. 
%For the example with non-metaphorical target word \textit{bother}, "\textit{Well, better than that, and one of the best things about it is the fact that so few people, except the French themselves, \underline{bother} with it.}", \textit{French} (which could also be understood as a language), \textit{bother}'s neighbour misleads baselines yet our model accurately detects its real subject \textit{people} here. %The above two examples from VUA are both classified successfully by our model but incorrectly by other denoising methods and MelBERT. 
%\textcolor{red}{[CL: where are those two examples from? VUA? MoH??]}\textcolor{blue}{[VUA]}

\begin{table*}[t]  \centering 
\small
% \resizebox{\columnwidth}{!}{
    \begin{tabular}{l|ccc|ccc|c|c|c} 
    \toprule
       Sent.   & \multicolumn{3}{c|}{RoPPT}                    & \multicolumn{3}{c|}{MelBERT*}  & F1 &  Pruning  & \# of  \\ \cline{2-7}
    len.   & Prec          & Rec           & F1             & Prec  & Rec   & F1      & diff.& comp. & Sent. \\ \hline
    <20  &76.4      & 74.8     & \textbf{75.6}              & 75.0  & 75.2  & 75.1  & \textbf{0.5}    & 10.7 / 12.3   & 18,515\\
    20-40  & 81.8   & 79.9      & \textbf{80.8}     & 79.2  & 79.1  & 79.2      & \textbf{1.6}        & 16.4 / 29.4   & 17,729\\ 
    >40  & 82.3     & 80.0      & \textbf{81.1}     & 78.5  & 76.8  & 77.6      & \textbf{3.5}        & 19.5 / 53.6    & 7,703\\ \bottomrule 
    \end{tabular}
% }
\caption{RoPPT and MelBERT* performance comparison on sentences with different length range from VUA18. `Pruning comp.' is the comparison of the average length of (pruned) / (original) sentences.}
\label{table:Length_Study}
\end{table*}

\section{Conclusion}

In this paper, we proposed, RoPPT, an effective approach to extract contextual information for target words for metaphor detection based on a target-oriented parse tree structure. Extensive experiments show that our model can yield better performance compared to the state-of-the-art. In addition, our method is particularly effective in denoising long sentences, despite its simplicity. 

\section{Limitation}
Empirical experiments show that  our method is more effective in denoising long sentences with the proposed target-oriented parse tree. While this is somewhat expected as shorter sentences tend to have cleaner context, it raises a question or limitation of how can we improve the proposed method to better deal with short sentences and improve its performance in these cases. One possibility is to exploit external knowledge (e.g. ConceptNet) to support the detection of the most important contextual words. 
%Our method mainly focuses on denoising long sentences with extracting main structure by utilizing a target-oriented parse tree, which might have less efficiency when working on short sentences. However, for short samples, we consider that it's semantic extracting and disambiguation which are much more important than denoising.

% Entries for the entire Anthology, followed by custom entries
\bibliography{anthology,custom, reference}
\bibliographystyle{acl_natbib}
% \\
% \\

% \appendix
%\section{Appendices}
%\label{sec:appendix}

%\begin{table}[!h]
%\resizebox{\columnwidth}{!}{
%\centering
%\begin{tabular}{lc}
%\toprule
 % & TITAN RTX  \\ \midrule
 %Runtime per epoch & 50 min  \\ 
 %Parameters  & 252,839,426 \\
 % \bottomrule
%\end{tabular}
%}
%\caption{Hardware and runtime details.}  
%\label{tabel:experiment}
%\end{table}

\end{document}